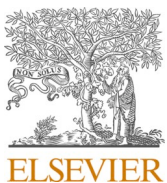
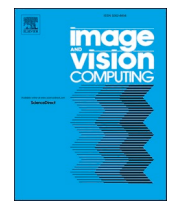

# ASF-YOLO: A novel YOLO model with attentional scale sequence fusion for cell instance segmentation

Ming Kang, Chee-Ming Ting [*], Fung Fung Ting, Raphaël C.-W. Phan

*School of Information Technology, Monash University, Malaysia Campus, Subang Jaya 47500, Malaysia*

A B S T R A C T

We propose a novel Attentional Scale Sequence Fusion based You Only Look Once (YOLO) framework (ASF-YOLO) which combines spatial and scale features for accurate and fast cell instance segmentation. Built on the YOLO segmentation framework, we employ the Scale Sequence Feature Fusion (SSFF) module to enhance the multiscale information extraction capability of the network, and the Triple Feature Encoder (TFE) module to fuse feature maps of different scales to increase detailed information. We further introduce a Channel and Position Attention Mechanism (CPAM) to integrate both the SSFF and TFE modules, which focus on informative channels and spatial position-related small objects for improved detection and segmentation performance. Experimental validations on two cell datasets show remarkable segmentation accuracy and speed of the proposed ASF-YOLO model. It achieves a box mAP of 0.91, mask mAP of 0.887, and an inference speed of 47.3 FPS on the 2018 Data Science Bowl dataset, outperforming the state-of-the-art methods. The source code is available at https://github.com/mkang315/ASF-YOLO.

## 1. Introduction

With the rapid development of sample preparation technology and microscopic imaging technology, quantitative processing and analysis of cell images play an important role in fields such as medicine and cell biology. Based on Convolutional Neural Networks (CNN), the characteristic information of different cell images can be learned through neural network training, which has strong generalization performance. The two-stage R-CNN series [1–3] and its one-stage variants [4,5] are classical CNN-based frameworks that have been used for segmentation tasks. However, these traditional methods based on CNNs only achieved sub-optimal performance for real-time cell instance segmentation, especially when dealing with dense and small cells.

In recent works, the You Only Look Once (YOLO) series [6–9] have become among the fastest and most accurate models for real-time instance segmentation. Because of the one-stage design idea and the capabilities of feature extraction, YOLO instance segmentation models have better accuracy and speed than two-stage segmentation models. Nevertheless, the performance of YOLO-based models for small object segmentation in medical or histopathology images, such as cell instance segmentation is largely unexplored. Cell instance segmentation poses more challenges due to the small, dense, and overlapping objects, as well as blurred boundaries of cells, which may result in poor segmentation accuracy. It requires accurate detailed segmentation of different types of objects in cell images. As shown in Fig. 1, different types of cell images have large differences in color, morphology, texture, and other characteristic information due to differences in cell morphology, preparation methods, and imaging technologies. Despite its improved segmentation accuracy and speed for natural images, the architectures of YOLO-based models can be further optimized for handling small objects in medical images, such as cells.

The typical YOLO framework architecture consists of three main parts: backbone, neck, and head. The backbone network of YOLO is a convolutional neural network that extracts image features at different granularities. Cross Stage Partial [10] Darknet with 53 convolutional layers (CSPDarknet53) [11] was modified from YOLOv4 [12] and designed as the backbone network of YOLOv5 [8], which contains C3 (CSP bottleneck including 3 convolutional layers) and ConvBNSiLU modules. The C3 modules are replaced by the C2f (CSP bottleneck including 2 convolutional layers with shortcut) modules in the backbone of YOLOv8 [9], which is the only difference from that of YOLOv5. As shown in Fig. 2, the level 1–5 feature extraction branches {P1, P2, P3, P4, P5} in the backbone of YOLOv5 and YOLOv8 correspond to the outputs of the YOLO network associated with each of these feature maps.

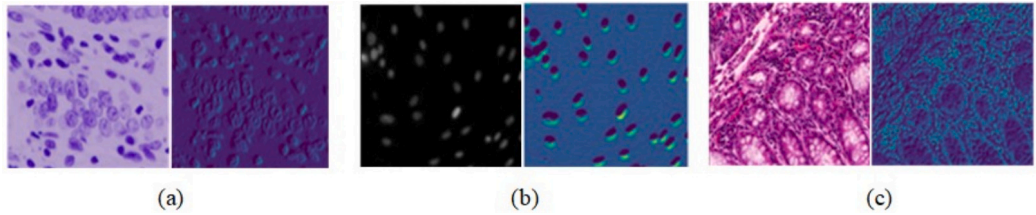

**Fig. 1.** Different cell images (left) and their feature maps (right).

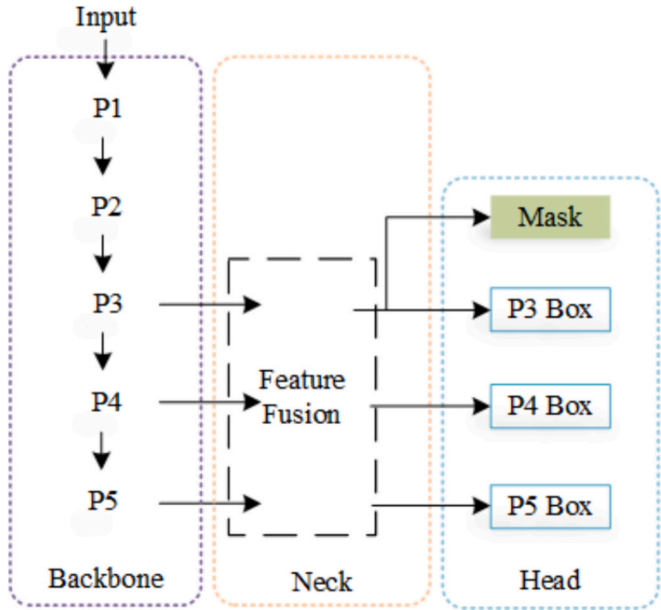

**Fig. 2.** An abridged general view of the framework of YOLOv5 v7.0 and YOLOv8 for segmentation task. P1, P2, P3, P4, P5 represent different levels of features output by the backbone. The head part clips the segmentation masks to bind them inside each of the detected bounding boxes, which ensures that the segmentation masks do not flow out of the bounding boxes. The neck part is intentionally abridged due to the different structures between YOLOv5 and YOLOv8.

YOLOv5 and YOLOv8 are the first mainstream YOLO-based architectures that can handle segmentation tasks besides detection and classification. In the feature extraction stage of YOLOv5, the CSPDarkNet53 backbone network stacked by multiple C3 modules is used, and then the three effective feature branches P3, P4, and P5 of the backbone network are used as the input of the Feature Pyramid Network (FPN) [13] structure to build multiscale fusion structure in the neck part. During the decoding process of the feature layer, three heads of different sizes corresponding to the effective feature branch of the backbone network are used for the bounding box prediction of the object. After upsampling the P3 features, pixel-by-pixel decoding is performed as the segmentation mask prediction of the object to complete the instance segmentation of the object. In the segmentation head, three scales of features output three different anchor boxes, and the mask proto module is responsible for outputting the prototype masks, which are processed to get the detection boxes and segmentation masks, for instance, segmentation tasks.

In this paper, we address the challenges of small object segmentation based on an improved YOLO-based model with an application to cell instance segmentation. Our method builds on the YOLOv5 model, leveraging the capability of its backbone in extracting multiscale features from the cell images. Our method further extends the original YOLOv5 architecture to improve its effectiveness in segmenting small objects by incorporating several new modules, especially in the neck part of the model. These include the fusion of multiscale features and the incorporation of attention mechanism.

More precisely, we propose a one-stage instance segmentation model for cell images, which integrates Attentional Scale Sequence Fusion in the YOLO framework (ASF-YOLO). The CSPDarknet53 backbone network is first used to extract multi-dimensional feature information from cell images in the feature extraction stage. We propose several novel network designs in the neck part to enable multiscale feature fusion and attention mechanism. The multiscale fusion aims to improve the models robustness to the scale variations of small objects from cell images obtained from different conditions. The latter provides a selective focus on the multiscale features relevant to the small objects. For the detection head, we leverage the EIoU [14] in the training stage to optimize the bounding box location loss by minimizing the difference between the width and height of the bounding box and the anchor box. The use of EIoU can better capture the locations of small objects, compared to CIoU used in YOLOv5 and YOLOv8, which only reflects the difference in aspect ratio, not the true relationship between the width and height of the labeled and predicted boxes. Soft Non-Maximum Suppression (Soft-NMS) [15] is also used in the post-processing stage to improve the densely overlapping cell problem. The main contributions of this work are summarized as follows:

1) We design a Scale Sequence Feature Fusion (SSFF) module and a Triple Feature Encoder (TFE) module to fuse the multiscale feature maps extracted from the backbone in a Path Aggregation Network (PANet) [16] structure. The SSFF combines global semantic information of images across different scales by normalizing, upsampling, and concatenating multiscale features into a 3D convolution. Thus, it can effectively handle objects of varying sizes, orientations, and aspect ratios in a scale-space representation to improve object segmentation. The TFE incorporates small, medium, and large-sized feature maps to capture the fine spatial information of small objects across distinctive scales. These overcome the limitations of FPN in YOLOv5, which cannot fully exploit the correlations between the pyramid feature maps via simple sum and concatenation operations and mainly leverages small feature maps.

2) We then design a Channel and Position Attention Mechanism (CPAM) to integrate feature information from the SSFF and TFE modules. This module allows the model to adaptively adjust its focus on relevant channels and spatial locations relevant for the small objects at different scales, and hence better instance segmentation than the conventional YOLOv5 architecture with no attention mechanism.

3) We apply the proposed ASF-YOLO model for challenging instance segmentation tasks of densely overlapping and various cell types. To our best knowledge, this is the first work to leverage a YOLO-based model for cell instance segmentation. Evaluation of two benchmarking cell datasets shows superior detection accuracy and speed compared to other state-of-the-art methods, including CNN-based models previously used for cell segmentation and several recent YOLO-based models.

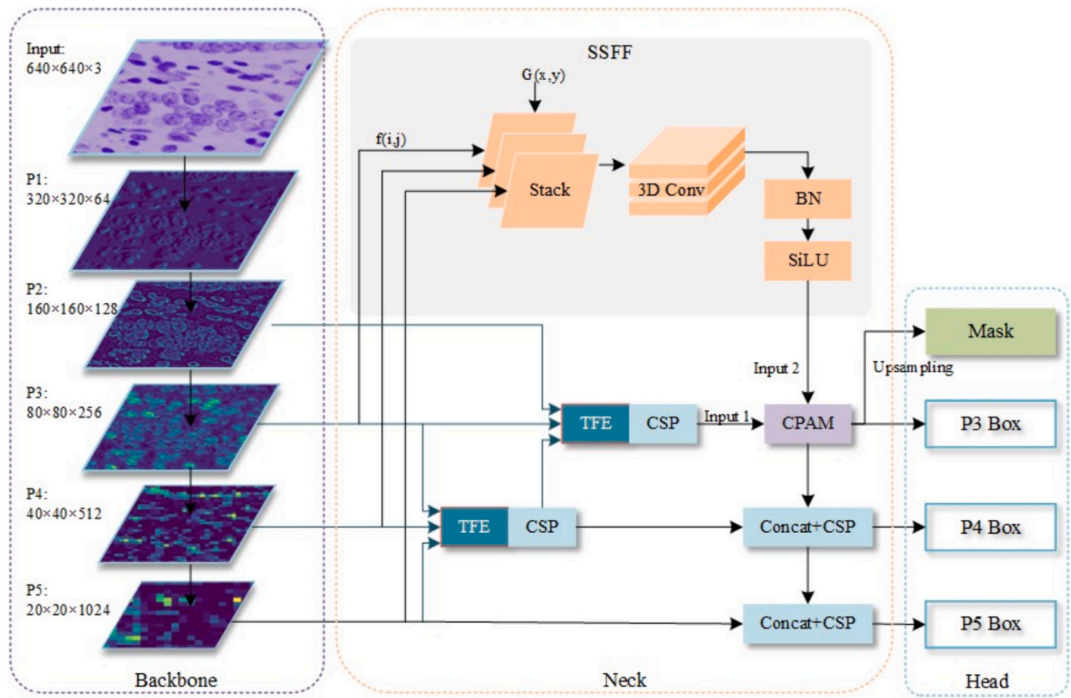

**Fig. 3.** The overview of the proposed ASF-YOLO model. The framework is mainly comprised of the Scale Sequence Feature Fusion (SSFF) module, the Triple Feature Encoder (TFE) module, and the Channel and Position Attention Model (CPAM) based on the CSPDarkNet backbone and the YOLO head. CSP and Concat modules come from YOLOv5.

## 2. Related work

### 2.1. Cell instance segmentation

Cell instance segmentation can further help complete the cell counting task in the image while semantic segmentation of cell images cannot. Deep learning approaches have increased the accuracy of automated nucleus segmentation [17]. Johnson et al. [18], Jung et al. [19], Fujita et al. [20] and Bancher et al. [21] proposed improved methods for simultaneous detection and segmentation of cells based on Mask R-CNN [2]. Yi et al. [22] and Cheng et al. [23] utilized a Single-Shot multi-box Detector (SSD) [24] method to detect and segment neural cell instances. Mahbod et al. [25] employed a semantic segmentation algorithm U-Net [26] based model for cell nuclei segmentation. The hybrid model SSD and U-Net with attention mechanisms [19] or U-Net and Mask R-CNN [27] achieved some boost in performance on the cell instance segmentation datasets. BlendMask [28] is a nuclei instance segmentation framework with a dilated convolution aggregation module and a context information aggregation module. Mask R-CNN is a two-stage object segmentation framework, of which speed is slow. SSD, U-Net, and BlendMask are unified end-to-end (i.e., one-stage) frameworks but have poor performance in segmenting dense and small cells. Traditional CNN-based methods for cell instance segmentation no longer meet the needs of real-time detection and segmentation.

### 2.2. Improved YOLO for instance segmentation

Recent improvements of YOLO, for the instance segmentation task, focus on attention mechanisms, improved backbone or networks, and loss functions. Squeeze-and-Excitation (SENet) [29] block was integrated into an improved YOLACT [6] for identifying rumen protozoa in microscopic images [30]. YOLOMask [31], PR-YOLO [32] and YOLO-SF [33] augmented YOLOv5 [8] and YOLOv7-Tiny [34] with Convolutional Block Attention Module (CBAM) [35]. Effective feature extraction modules were added to the improved backbone network to make the process of YOLO feature extraction more efficient [36,37]. YOLO-CORE [38] enhanced the mask of the instance efficiently by explicit and direct contour regression using a designed multi-order constraint consisting of a polar distance loss and a sector loss. In addition, the hybrid models YOLOMask [39] and YUSEG [40] combined optimized YOLOv4 [12] and the original YOLOv5s with semantic segmentation U-Net network to ensure the accuracy of instance segmentation. To our best knowledge, these improved YOLO architectures, originally designed for instance segmentation for natural images, have not been applied for cell instance segmentation, which is more challenging due to the small and densely overlapping cells.

## 3. The proposed ASF-YOLO model

### 3.1. Overall architecture

Fig. 3 shows the overview of the proposed ASF-YOLO framework that combines spatial and multiscale features for cell image instance segmentation. We develop a novel feature fusion network architecture consisting of two main component networks that can provide complementary information for small object segmentation: (1) SSFF module, which combines global or high-level semantic information from multiple scales of images, and (2) TFE module, which can capture local fine details of small objects. The integration of both local and global feature information can produce a more accurate segmentation map. We perform a fusion of output features of P3, P4, and P5 extracted from the backbone network. First, the SSFF module is designed to effectively fuse the feature maps of P3, P4, and P5 which captures the different spatial scales covering a variety of sizes and shapes of different cell types. In SSFF, the P3, P4, and P5 feature maps are normalized to the same size, upsampled, and then stacked together as input to a three-dimensional (3D) convolution to combine multiscale features. Secondly, the TFE module is developed to enhance small object detection for dense cells, by splicing features of three different sizes—large, medium, and small-—in the spatial dimension to capture detailed information about small objects. The detailed feature information of the TFE module is then integrated into each feature branch through the PANet structure, which is then combined with the multiscale information of the SSFF module into the P3 branch. We further introduce the Channel and Position Attention Mechanism (CPAM) in the P3 branch to leverage both the high-level multiscale features and the detailed features. The channel and position attention mechanism in the CPAM can respectively capture informative channels and refine spatial localization related to small objects like cells, thus enhancing its detection and segmentation accuracy.

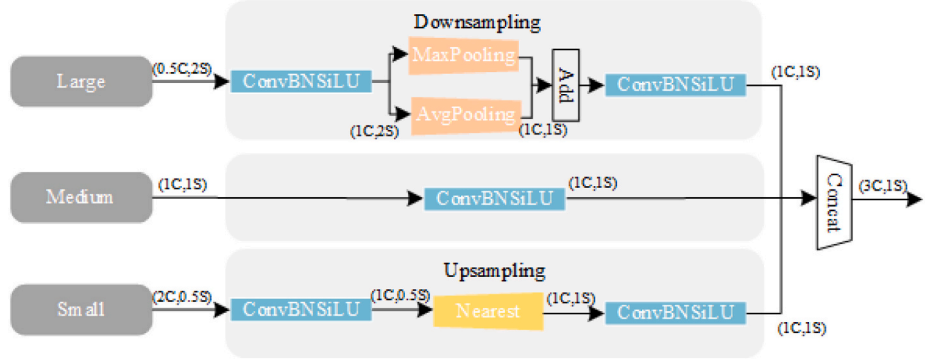

**Fig. 4.** The structure of TFE module. *C* represents the number of channels, and *S* represents the feature map size. Each triple feature encoder module uses three feature maps of different sizes as input.

*3.2. Scale sequence feature fusion module*

For the multiscale problem of cell images, feature pyramid structures are used for feature fusion in the existing literature, in which only sum or concatenation is employed to fuse the pyramid features. However, the structures of various feature pyramid networks cannot effectively exploit the correlation between all pyramid feature maps. We propose a novel SSFF module that can better combine the multiscale feature maps, i.e., the high-level information of deep feature maps with the detailed information of shallow feature maps, which have the same aspect ratio.

We further construct the sequential representations of the multiscale feature maps generated from the backbone (i.e., P3, P4, and P5) that capture the image content at different levels of detail or scale. The feature maps P3, P4, and P5 are first convolved with a series of Gaussian kernels of increasing standard deviation [41–43], as follows:

$$F_\sigma(i,j) = \sum_u \sum_v f(i-u, i-v) \times G_\sigma(u,v) \quad (1)$$

$$G_\sigma(x,y) = \frac{1}{2\pi\sigma^2} e^{-(x^2+y^2)/2\sigma^2} \quad (2)$$

where $f$ represents a two-dimensional (2D) feature map and $F_\sigma$ is generated by smoothing with a series of convolutions using a 2D Gaussian filter with increasing standard deviation $\sigma$.

Then, we stack these feature maps of different scales horizontally and use 3D convolution to extract their scale sequence features, inspired by 2D and 3D convolution operations on the multiple video frames [44]. Since the output feature maps from the above Gaussian smoothing have different resolutions, the nearest neighbor interpolation method is used to align all feature maps to the same resolution as the P3. This is because the high-resolution feature map level P3 contains most of the information crucial for the detection and segmentation of small objects, the SSFF module is designed based on the P3 level. As shown in Fig. 3, the proposed SSFF module consists of the following components:

- A 1 × 1 convolution is used to change the number of channels of the P4 and P5 feature levels to 256.
- Nearest neighbor interpolation method [45] is used to adjust their size to the size of the P3 level.

- The unsqueeze method is used to increase the dimension of each feature layer, changing it from a 3D tensor [height, width, channel] to a 4D tensor [depth, height, width, channel].
- The 4D feature maps are then concatenated along the depth dimension to form a 3D feature map for subsequent convolutions.
- Finally, 3D convolution, 3D batch normalization, and SiLU [46] activation function are used to complete scale sequence feature extraction.

*3.3. Triple feature encoding module*

To identify densely overlapping small objects, one can reference and compare shape or appearance changes at different scales by enlarging the image. Since different feature layers of the backbone network have different sizes, the conventional FPN fusion mechanism only upsamples the small-sized feature map and then splits or adds it to the features of the previous layer, ignoring the rich detailed information of the larger-sized feature layer. Therefore, we propose the TFE module, which splits large, medium, and small features, adds large-size feature maps, and performs feature amplification to improve detailed feature information.

Fig. 4 illustrates the structure of the TFE module. Before feature encoding, the number of feature channels is first adjusted so that it is consistent with the main scale characteristics. After the large-size feature map (Large) is processed by the convolution module, its channel number is adjusted to 1C. Then, a hybrid structure of maximum pooling + average pooling is used for downsampling, which helps to reduce the spatial dimensions of features and to achieve translation invariance, enhancing the network's robustness to spatial variations and translations of the input images. For small-size feature maps (Small), the convolution module is also used to adjust the number of channels, and then the nearest neighbor interpolation method is used for upsampling. This helps to retain the local features and prevents the loss of small object feature information because the nearest-neighbor interpolation technique can populate the feature map by utilizing neighboring pixels and accounts for sub-pixel neighborhoods. Additionally, when employing nearest-neighbor interpolation for upsampling, a significant portion of the feature details pertaining to small objects tend to be lost due to background interference. Finally, the three feature maps of large, medium, and small sizes with the same dimensions are convolved once and then spliced in the channel dimension, as follows:

$$F_{TFE} = Concat(F_l, F_m, F_s) \quad (3)$$

where $F_{TFE}$ denotes the feature maps of the TFE module output. $F_l$, $F_m$, and $F_s$ denote large, medium, and small-size feature maps, respectively. $F_{TFE}$ results from the concatenation of $F_l$, $F_m$, and $F_s$. $F_{TFE}$ has the same resolution as and three times the channel number of $F_m$.

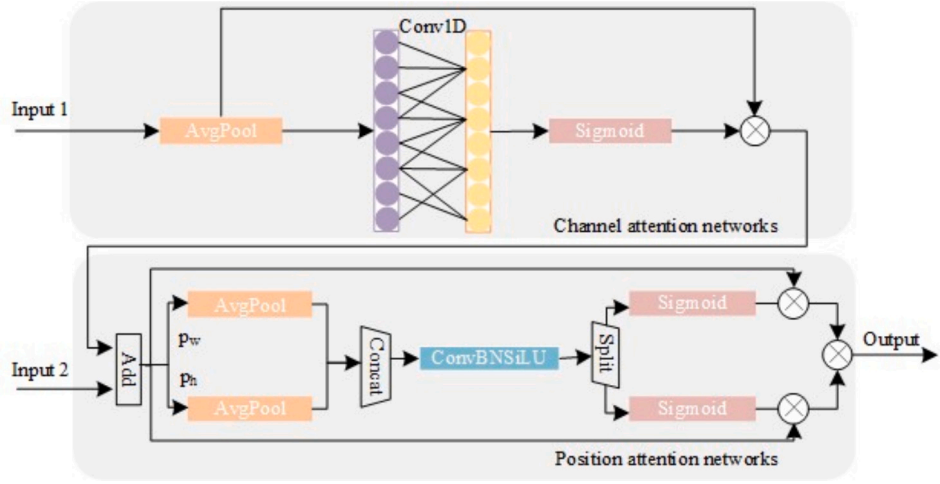

**Fig. 5.** The structure of CPAM module. It contains channel and position attention networks. *w* and *h* represent width and height, respectively. ⊗ denotes the operation of the Hadamard product.

*3.4. Channel and position attention mechanism*

To extract the representative feature information contained in different channels, we propose the CPAM integrate detailed and multi-scale feature information from both SSFF and TFE. The architecture of CPAM is shown in Fig. 5. It consists of a channel attentional network receiving input from the TFE (Input 1), and a position attentional network receiving input from the superposition of the outputs of the channel attentional network and the SSFF (Input 2).

Input 1 for the channel attentional network is the feature map after PANet, which contains the detailed features of the TFE. The SENet [29] channel attention block first adopted global average pooling for each channel independently and used two fully connected layers together with a nonlinear Sigmoid function to generate channel weights. The two fully connected layers aim to capture nonlinear cross-channel interactions, which involves reducing dimensionality to control the complexity of the model, but dimensionality reduction brings side effects to channel attention prediction, and capturing the dependencies between all channels is inefficient and unnecessary. We introduce an attention mechanism without dimensionality reduction to capture cross-channel interactions in an effective manner. After channel-wise global average pooling without reducing dimensionality, local cross-channel interactions are captured by considering each channel and its $k$ nearest neighbors. This is implemented using $1D$ convolutions of size $k$, where the kernel size $k$ represents the coverage of local cross-channel interactions, that is, how many neighbors participate in the attention prediction of one channel. To obtain optimal coverage, one may resort to manual tuning of $k$ in different network structures and different numbers of convolution modules, which is tedious. Since the convolution kernel size $k$ is proportional to the channel dimension $C$, the channel dimension is generally an exponent of 2 and can defined in terms of $k$ as

$$C = \psi(k) = 2^{(\gamma \times k - b)} \tag{4}$$

where $\gamma$ and $b$ are the scaling parameters controlling the ratio of the convolution kernel size $k$ to the channel dimension $C$, respectively.

$$k = \Psi(C) = \left| \frac{\log_2(C) + b}{\gamma} \right|_{odd} \tag{5}$$

where $|\cdot|_{odd}$ denotes the odd number of the nearest neighbors. The value of $\gamma$ is set to 2 and $b$ is set to 1. According to the above non-linear mapping relationship, the exchange of high-value channels is longer, while the exchange of low-value channels is shorter. Therefore, the channel attention mechanism can perform deeper mining of multiple channel features.

Combining the outputs of the channel attention mechanism with the features from SSFF (Input 2) as input to the position attention network provides complementary information to extract crucial location information from each cell. In contrast to the channel attention mechanism, the position attention mechanism first splits the input feature map into two parts in terms of its width and height, which are then processed separately for feature encoding in the axes ($p_w$ and $p_h$), and are finally merged to generate output.

More precisely, the input feature map is pooled in both horizontal ($p_w$) and vertical ($p_h$) axes to retain the spatial structure information of the feature map, which can be calculated as follows:

$$p_w(i) = \frac{1}{H}\sum_{0 \leq j \leq H} E(i,j) \tag{6}$$

$$p_h(j) = \frac{1}{W}\sum_{0 \leq i \leq W} E(i,j) \tag{7}$$

where $W$ and $H$ are the width and height of the input feature map, respectively. $E(i,j)$ are the values at the position $(i,j)$ of the input feature map.

When generating position attention coordinates, the concatenation and convolution operations are applied for the horizontal and vertical axes:

$$P(a_w, a_h) = Conv[Concat(p_w, p_h)] \tag{8}$$

where $P(a_w, a_h)$ denotes the output of position attention coordinates, *Conv* denotes a $1 \times 1$ convolution and *Concat* denotes concatenation.

When splitting the attention features, pairs of location-dependent feature maps are generated as follows:

$$s_w = Split(a_w) \tag{9}$$

$$s_h = Split(a_h) \tag{10}$$

where $s_w$ and $s_h$ are the width and height of the output of the splitting, respectively.

The final output of CPAM is defined by:

$$F_{CPAM} = E \times s_w \times s_h \tag{11}$$

where $E$ represents the weight matrix of the channel and position attentions.

## 3.5. Anchor box optimization

By optimizing the loss function and Non-Maximum Suppression (NMS), the anchor boxes of the three detection heads are improved for a better instance segmentation of cell images in different sizes.

Intersection over Union (IoU) is typically used as the anchor box loss function to determine convergence by calculating the degree of overlap between the labeled bounding box and the prediction box. However, the classical IoU loss cannot reflect the distance and overlap between the object box and the anchor box. To address these issues, GIoU [47], DIoU, and CIoU [48] have been proposed. CIoU introduces an influence factor based on DIoU Loss, which is used by YOLOv5 and YOLOv8. While taking into account the impact of the overlapping area and center point distance on the loss function, it also takes into account the impact of the width-to-height (i.e., aspect) ratio of the labeled box and the predicted box on the loss function. However, it only reflects the difference in aspect ratio, rather than the true relationship between the width and height of the labeled box and the predicted box. EIoU [14] minimizes the difference in width and height between the object box and anchor box, which can improve the location effect of small objects. EIoU loss can be divided into 3 parts: the IoU loss function $L_{IoU}$, distance loss function $L_{dis}$, and aspect loss function $L_{asp}$, of which the formula is as follows.

$$L_{EIoU} = L_{IoU} + L_{dis} + L_{asp} = 1 - IoU + \frac{\rho^2(b, b_{gt})}{w_c^2 + h_c^2} + \frac{\rho^2(w, w_{gt})}{w_c^2} + \frac{\rho^2(h, h_{gt})}{h_c^2} \quad (12)$$

where $\rho(\cdot) = \|b - b_{gt}\|_2$ indicates the Euclidean distance and $b$ and $b_{gt}$ denote the central points of $B$ and $B_{gt}$ respectively; $b_{gt}$, $w_{gt}$, and $h_{gt}$ are the central point b, width, and height of ground truth; $w_c$ and $h_c$ denote the width and height of the smallest enclosing box covering the two boxes. Compared with CIoU, EIoU not only speeds up the convergence speed of the prediction frame but also improves the regression accuracy. Therefore, we select EIoU to replace CIoU in the head part.

In order to eliminate duplicate anchor boxes, the detection models output multiple detection boundaries at the same time, especially when there are many high-confidence detection boundaries around the real objects. The principle of the classical NMS [49] algorithm is to get the local maximum. If the difference between the current bounding box and the highest-scoring detection frame is greater than the threshold, the score of the bounding box is directly set to zero. To overcome the error caused by the classical NMS, We adopt Soft-NMS [15], which uses the Gaussian function as the weight function to reduce the score of the prediction boundaries to replace the original score instead of directly setting it to zero, thus modifying the rules of removing the bounding box.

## 4. Experiments

### 4.1. Datasets

We evaluated the performance of the proposed ASF-YOLO model on two cell image datasets: DSB2018 and BCC datasets. The 2018 Data Science Bowl (DSB2018) dataset [50] contains 670 cell nuclei images with segmented masks, which is designed to assess the generalizability of an algorithm across variations of the cell type, magnification, and imaging modality (brightfield vs. fluorescence). Each mask contains one nucleus, with no overlapping between masks (no pixel belongs to two masks). The dataset was randomly divided into the training set and the test set in terms of an 8:2 ratio. The sample size of the training set and test set are 536 and 134 images, respectively.

The Breast Cancer Cell (BCC) dataset [51] was collected from the Center for Bio-Image Informatics, University of California, Santa Barbara (UCSB CBI). It includes 160 hematoxylin and eosin-stained histopathology images used in breast cancer cell detection, with associated ground truth data. The dataset was randomly partitioned into 128 images (80%) as the training set, and 32 images (20%) as the test set.

**Table 1**
Performance comparison of different models for cell instance segmentation on the DSB2018 dataset. The best results are in bold.

| Model | Param (M) | Box | | Mask | | |
|---|---|---|---|---|---|---|
| | | $mAP_{50}$ | $mAP_{50:95}$ | $mAP_{50}$ | $mAP_{50:95}$ | FPS |
| Mask R-CNN [2] | 43.75 | 0.774 | 0.519 | 0.782 | 0.525 | 20 |
| Cascade Mask R-CNN [3] | 69.17 | – | – | 0.783 | 0.533 | 17.9 |
| SOLO [4] | – | – | – | 0.642 | 0.398 | 25 |
| SOLOv2 [5] | – | – | – | 0.741 | 0.495 | 28.7 |
| YOLACT [6] | 34.3 | 0.703 | 0.456 | 0.683 | 0.440 | 25 |
| Mask RCNN Swin T [52] | 47.3 | 0.784 | 0.524 | 0.783 | 0.527 | 24 |
| YOLOv5l-seg [8] | 45.27 | 0.876 | 0.616 | 0.855 | 0.502 | 46.9 |
| YOLOv8l-seg [9] | 45.91 | 0.865 | 0.631 | 0.866 | 0.562 | 45.5 |
| **ASF-YOLO (Ours)** | 46.18 | **0.910** | **0.676** | **0.887** | **0.558** | **47.3** |

**Table 2**
Performance comparison of different models for cell instance segmentation on the BCC dataset. The best results are in bold.

| Model | Box | | Mask | |
|---|---|---|---|---|
| | $mAP_{50}$ | $mAP_{50:95}$ | $mAP_{50}$ | $mAP_{50:95}$ |
| Mask R-CNN [2] | 0.852 | 0.614 | 0.836 | 0.628 |
| Cascade Mask R-CNN [3] | 0.836 | 0.630 | 0.823 | 0.598 |
| SOLO [4] | – | – | 0.864 | 0.647 |
| SOLOv2 [5] | – | – | 0.860 | 0.651 |
| YOLACT [6] | 0.715 | 0.545 | 0.774 | 0.565 |
| Mask RCNN Swin T [52] | 0.841 | 0.604 | 0.806 | 0.588 |
| YOLOv5l-seg [8] | 0.892 | 0.703 | 0.877 | 0.672 |
| YOLOv8l-seg [9] | 0.850 | 0.619 | 0.814 | 0.564 |
| **ASF-YOLO (Ours)** | **0.911** | **0.737** | **0.898** | **0.645** |

### 4.2. Implementation details

The experiments were implemented on the NVIDIA GeForce 3090 (24G) GPU and Pytorch 1.10, Python 3.7, and CUDA 11.3 dependencies. We employed the initial weight of the pretrained COCO dataset. The input image size is $640 \times 640$. The batch size of the training data quantity is 16. The training process lasts 100 epochs. We used Stochastic Gradient Descent (SGD) as an optimization function to train the model. The hyperparameters of SDG are set to 0.9 of the momentum, 0.001 of the initial learning rate, and 0.0005 of the weight decay.

### 4.3. Quantitative results

Table 1 shows performance comparison on the DSB2018 dataset between the proposed ASF-YOLO with other classical and state-of-the-art methods including Mask R-CNN [2], Cascade Mask R-CNN [3], SOLO [4], SOLOv2 [5], YOLACT [6], Mask R-CNN with Swin Transformer backbone (Mask RCNN Swin T) [52], YOLOv5l-seg v7.0 [8], and YOLOv8l-seg [9].

Our model with 46.18 million parameters achieved the best accuracy with Box $mAP_{50}$ of 0.91 and Mask $mAP_{50}$ of 0.887 and the inference speed reached 47.3 Frame Per Second (FPS), which is the best performance. Due to the image input size of $800 \times 1200$, the accuracy and speed of Mask R-CNN using the Swin Transformer backbone are not high. Our model also surpasses the classical one-stage algorithms SOLO and YOLACT.

Our proposed model also achieved the best instance segmentation performance on the BCC dataset, as shown in Table 2. The experimental validation shows the generalization ability of ASF-YOLO to different datasets with varying cell types.

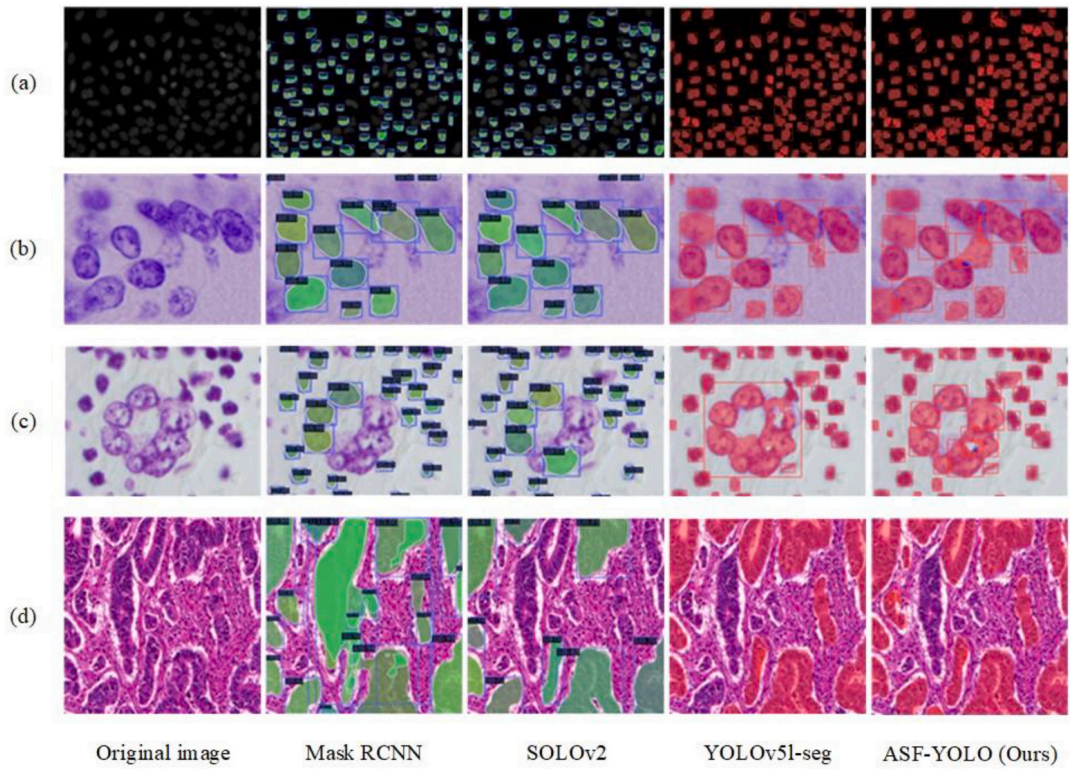

**Fig. 6.** Qualitative comparison of different instance segmentation models on the DSB2018 dataset.

**Table 3**
Ablation study of main components of the ASF-YOLO on the DSB2018 dataset.

| Method | | | | | Box | | Mask | |
|---|---|---|---|---|---|---|---|---|
| Soft-NMS | EIoU | TFE | SSFF | CPAM | $mAP_{50}$ | $mAP_{50:95}$ | $mAP_{50}$ | $mAP_{50:95}$ |
|  |  |  |  |  | 0.876 | 0.616 | 0.855 | 0.502 |
| ✓ |  |  |  |  | 0.881 | 0.622 | 0.856 | 0.523 |
|  |  | ✓ |  |  | 0.880 | 0.634 | 0.852 | 0.507 |
| ✓ | ✓ | ✓ |  |  | 0.891 | 0.653 | 0.867 | 0.542 |
| ✓ | ✓ |  | ✓ |  | 0.896 | 0.653 | 0.876 | 0.549 |
| ✓ | ✓ | ✓ | ✓ |  | 0.902 | 0.656 | 0.874 | 0.543 |
| ✓ | ✓ | ✓ | ✓ | ✓ | 0.910 | 0.676 | 0.887 | 0.558 |

## 4.4. Qualitative results

Fig. 6 provides a visual comparison of cell segmentation by different methods on sample images from the DSB2018 dataset. By using the TFE module to improve small object detection performance, ASF-YOLO achieved a good recall value for cell images with dense and small objects in a single channel. By using the SSFF module to enhance multiscale feature extraction performance, ASF-YOLO also provided a good segmentation accuracy for large-sized cell images under complex backgrounds. This indicates the good generalizability of our method to varying cell types. From Fig. 6 (a) and (b), each model has good results because the cell images are relatively simple. From Fig. 6 (c) and (d), Mask R-CNN has a high false detection rate due to the design principle of the two-stage algorithm. SOLO has many missed detections and YOLOv5l-seg fails to segment cells with blurred boundaries.

## 4.5. Ablation study

We conduct a series of extensive ablation studies of the proposed ASF-YOLO model.

**Table 4**
Effect of different attention mechanisms.

| Method | $mAP_{50}$ | | Param (K) | FLOPs (M) |
|---|---|---|---|---|
| ASF-YOLO w/o attention (baseline) | 0.902 | 0.874 | 0 | 0 |
| +SENet [29] | 0.901 | 0.879 | +8.19 | +1.65 |
| +CBAM [35] | 0.905 | 0.884 | +16.48 | +3.94 |
| +CA [53] | 0.903 | 0.8881 | +12.32 | +1.32 |
| +CPAM (ours) | 0.910 | 0.888 | +12.23 | +2.96 |

### 4.5.1. Effect of the proposed methods

Table 3 shows the contribution of each proposed module in improving the segmentation performance. The use of Soft-NMS in YOLOv5l-seg can overcome the error suppression problem due to mutual occlusion when detecting cells of dense small objects, and provide performance improvement. EIoU loss function improves the effect of small object bounding boxes, improving $mAP_{50:95}$ by 1.8%. The SSFF, TFE, and CPAM modules have effectively improved the model performance by solving small object instance segmentation of cell images.

### 4.5.2. Effect of attention mechanisms

Compared with the channel attention SENet, the channel and spatial attention CBAM, and the spatial attention Coordinate Attention (CA) [53], the proposed CPAM attention mechanism provides better performance despite with slight increase in computation amount and parameters, as shown in Table 4.

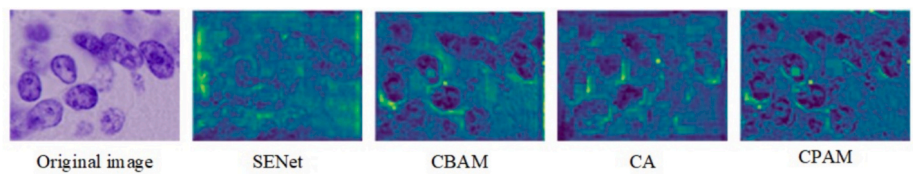

**Fig. 7.** Qualitative comparison of different attentional mechanisms, for instance, segmentation on the DSB2018 dataset.

**Table 5**
Effect of different convolution modules in the backbone of the proposed ASF-YOLO. The best results are in bold.

| Dataset | Module | Box | | Mask | |
|---|---|---|---|---|---|
| | | $mAP_{50}$ | $mAP_{50:95}$ | $mAP_{50}$ | $mAP_{50:95}$ |
| DSB2018 | C3 | **0.910** | **0.676** | **0.887** | 0.558 |
| | C2f | 0.867 | 0.633 | 0.859 | 0.558 |
| BCC | C3 | **0.911** | **0.737** | **0.898** | **0.645** |
| | C2f | 0.855 | 0.619 | 0.835 | 0.570 |

Fig. 7 shows the visualization of segmentation results using different attentional modules in the ASF-YOLO model. The proposed CPAM has better channel and positional feature information and mined richer features from the original images.

*4.5.3. Effect of convolution module in the backbone*

To demonstrate why we chose the YOLOv5 backbone network, we compare the results of the improved neck part with different YOLO backbone networks. Table 5 shows that, when the C3 modules of YOLOv5 are replaced by the C2f modules of YOLOv8 in the backbone of the proposed model, the performance of the C2f modules in the backbone decreases on the two datasets.

## 5. Conclusion

We developed an accurate and fast instance segmentation model ASF-YOLO for cell image analysis, which fuses spatial and scale features for the detection and segmentation of cell images. We introduced several novel modules in the YOLO framework. The SSFF and TFE modules enhance the multiscale and small object instance segmentation performance. The channel and position attention mechanism further mines the feature information of the two modules. Extensive experimental results demonstrate that our proposed model is capable of handling instance segmentation tasks for various cell images, and substantially improving the accuracy of the original YOLO models on cell segmentation due to small and dense objects. Our method substantially outperforms state-of-the-art methods in terms of both accuracy and inference speed for cell instance segmentation. Due to the small size of the data set in this article, the generalization performance of the model needs to be further improved. In addition, the effectiveness of each module in ASF-YOLO is discussed in the ablation study.

The proposed ASF-YOLO has achieved a good balance between detection accuracy and computational speed, despite the fact that the CPAM attention mechanism might induce a slight increase in computational effort. However, there is still room for further enhancing the overall detection accuracy while maintaining the model's segmentation efficiency, which is crucial for practical implementation in clinical settings. To this end, future work will extend the feature extractor to incorporate hierarchical convolutional structure, deformable convolution and non-local mechanisms as in [54] to enlarge the receptive field within the bottleneck block. Besides, the dilated convolution [55] could be used to enhance the performance of CNNs in capturing global contextual information without increasing the computational effort and the number of parameters. As inspired by [56], transfer learning could be performed on the backbone network to extract histological features. The recent advances in Transformer could also be adopted to improve the attention mechanism of the proposed model.

## CRediT authorship contribution statement

**Ming Kang:** Writing – review & editing, Writing – original draft, Visualization, Validation, Software, Investigation, Formal analysis, Conceptualization. **Chee-Ming Ting:** Writing – review & editing, Validation, Supervision, Project administration, Funding acquisition, Conceptualization. **Fung Fung Ting:** Writing – review & editing, Validation, Supervision. **Raphaël C.-W. Phan:** Writing – review & editing, Validation, Supervision.

## Declaration of competing interest

There are no competing interests to declare by the authors.

## Data availability

We have shared the link to our data in the References section.

## Acknowledgments

This work was supported by the Monash University Malaysia and the Ministry of Higher Education, Malaysia under Fundamental Research Grant Scheme FRGS/1/2023/ICT02/MUSM/02/1.

M. Kang et al.                                                                                                                                                                                            *Image and Vision Computing 147 (2024) 105057*